# A Hybrid CNN-Transformer Model for Heart Disease Prediction Using Life History Data

Ran Hao
University of North Carolina at Chapel Hill
Chapel Hill, USA

Yanlin Xiang
University of Houston
Houston, USA

Junliang Du
Shanghai Jiao Tong University
Shanghai, China

Qingyuan He
New York University
New York, USA

Jiacheng Hu
Tulane University
New Orleans, USA

Ting Xu*
University of Massachusetts Boston
Boston, USA

*Abstract*—This study proposed a hybrid model of a convolutional neural network (CNN) and a Transformer to predict and diagnose heart disease. Based on CNN's strength in detecting local features and the Transformer's high capacity in sensing global relations, the model is able to successfully detect risk factors of heart disease from high-dimensional life history data. Experimental results show that the proposed model outperforms traditional benchmark models like support vector machine (SVM), convolutional neural network (CNN), and long short-term memory network (LSTM) on several measures like accuracy, precision, and recall. This demonstrates its strong ability to deal with multi-dimensional and unstructured data. In order to verify the effectiveness of the model, experiments removing certain parts were carried out, and the results of the experiments showed that it is important to use both CNN and Transformer modules in enhancing the model. This paper also discusses the incorporation of additional features and approaches in future studies to enhance the model's performance and enable it to operate effectively in diverse conditions. This study presents novel insights and methods for predicting heart disease using machine learning, with numerous potential applications-especially in personalized medicine and health management.

## I. INTRODUCTION

Heart disease is one of the top killer diseases in the world. Because of this, doctors and researchers have always been on the lookout for ways to detect heart disease as early as possible. Recently, with the ongoing growth of large medical and health data, methods for detecting heart disease using artificial intelligence have started to surface. The early symptoms of heart disease are not typically easy to identify, and general diagnosis methods rely on experienced doctors' expertise, which may lead to missed or incorrect diagnoses [1,2]. Therefore, utilizing machine learning and deep learning methods for the detection and diagnosis of heart disease at an early stage can effectively improve the accuracy and efficiency of the diagnosis. Especially for those patients with no obvious symptoms, data-oriented models can provide more accurate auxiliary decision-making for physicians through handling their life history data, health monitoring data, etc[3,4].

This paper proposes a heart disease detection algorithm based on a convolutional neural network (CNN) and a Transformer model. The algorithm combines the local feature extraction capability of CNN and the advantages of Transformer in sequence modeling, aiming to extract potential health risk signals from patients' life history data [5]. Life history data usually includes a variety of information, such as eating habits, exercise frequency, sleep quality, weight changes, etc., which may have a profound impact on heart health in the long run. By deeply analyzing these multi-dimensional data, potential health risks can be better captured, and the possibility of heart disease in individuals in the future can be predicted. Therefore, how to effectively mine valuable features from these data is the key to improving the performance of heart disease detection algorithms.

In this study, we focus on the feature extraction and modeling process of life history data. Life history data itself has time series characteristics and often contains a variety of complex factors. Therefore, traditional machine learning methods may have problems such as insufficient feature extraction and information loss when processing such data [6]. To this end, this paper uses a convolutional neural network (CNN) as the basic tool for feature extraction and uses its advantages in image data processing to extract meaningful local features from life history data. CNN can effectively learn the spatial relationship in the input data through multi-layer convolution operations, which is crucial for processing multi-dimensional and complex life data. At the same time, the Transformer model shows excellent performance in processing time series data and can effectively capture long-term dependencies and global information through the self-attention mechanism. Combining the advantages of CNN and Transformers, it is possible to more comprehensively and deeply analyze the complex features required for heart disease detection.

## II. RELATED WORK

The increasing availability of high-dimensional data, particularly in health and lifestyle monitoring, has driven extensive research into machine learning and deep learning techniques capable of extracting meaningful patterns from complex data sources.

Convolutional neural networks (CNNs) are widely recognized for their strong ability to extract local spatial patterns from multi-dimensional data. Xiao et al. [7] and Wang et al. [8] applied CNN-based models to high-dimensional datasets, demonstrating how hierarchical convolutional layers automatically learn localized features that are difficult to extract through manual feature engineering. While CNNs excel at capturing local dependencies, they struggle to model long-term dependencies and complex global relationships. Transformer models, with their self-attention mechanism, have emerged as a highly effective solution for learning such global patterns. Zhu et al. [9] demonstrated the ability of Transformers to capture long-range dependencies across sequential data, providing a clear example of how attention-based architectures can adaptively model interactions across different time points. Wu et al. [10] further enhanced this approach by incorporating adaptive attention and feature embedding techniques, allowing the model to dynamically focus on the most informative parts of the input. The combination of local and global feature extraction capabilities has been increasingly explored through hybrid deep learning architectures that integrate CNNs with Transformers. Gao et al. [11] proposed a hybrid model that combines transfer learning and meta-learning, demonstrating how combining complementary architectures can improve generalization, especially when data is sparse or noisy.

In addition to architectural advancements, effective feature selection and pattern discovery techniques play an important role in maximizing the predictive value of high-dimensional data. Li [12] proposed methods for identifying informative patterns from high-dimensional data while reducing redundancy, offering strategies that can be applied during pre-processing to select the most relevant features from life history records. Gao et al. [13] applied a hypergraph-enhanced sequential modeling framework to predict future visits in sequential data, demonstrating the value of modeling higher-order relationships across sequences. While this work does not explicitly adopt hypergraph-based techniques, the emphasis on preserving complex temporal and relational structures aligns with the core motivation for integrating CNNs and Transformers to comprehensively capture health-related signals from life history data. Similarly, Li et al. [14] introduced a matrix logic-based framework for efficient discovery of frequently occurring patterns, which could be adapted to identify recurring lifestyle behaviors linked to elevated heart disease risk. These techniques contribute indirectly to this work by informing strategies for data cleaning and feature selection, which are essential when dealing with complex and noisy life history data. Beyond individual architectures, capturing the structural relationships between events and features can further improve predictive performance. Yan et al. [15] applied neural networks to survival prediction across diverse conditions, demonstrating the adaptability of neural architectures to heterogeneous data sources, which aligns with this work's goal of creating a flexible and generalizable framework for predicting heart disease across diverse population subgroups. Wang [16] explored dynamic scheduling strategies for optimizing computing environments, providing insights into how system-level optimization can enhance the performance and scalability of machine learning models applied to large-scale health data analysis.

Further, adaptive learning and optimization techniques can enhance the robustness of predictive models when applied to diverse populations with varying lifestyle and health patterns. Huang et al. [17] introduced a reinforcement learning-based approach for dynamically optimizing data mining strategies in evolving environments, demonstrating how adaptive learning techniques can improve model performance when data distributions shift over time.

Finally, multi-source data fusion methods offer additional strategies for combining complementary information from various data streams, enriching the predictive power of disease risk models. Mei et al. [18] proposed collaborative hypergraph networks to enhance disease risk prediction, emphasizing the importance of combining heterogeneous data sources into a unified predictive framework. This multi-source fusion perspective further supports the rationale for integrating CNN and Transformer modules, where spatial, temporal, and contextual features extracted from different aspects of life history data contribute collectively to heart disease prediction. Taken together, these prior works provide the methodological foundation for the CNN-Transformer hybrid model proposed in this paper.

## III. METHOD

In this study, we proposed a heart disease detection algorithm based on convolutional neural network (CNN) and Transformer. The algorithm mainly relies on CNN to extract local features in life history data and models the global dependencies of data through the self-attention mechanism of Transformer, so as to accurately predict heart disease. Its model architecture is shown in Figure 1.

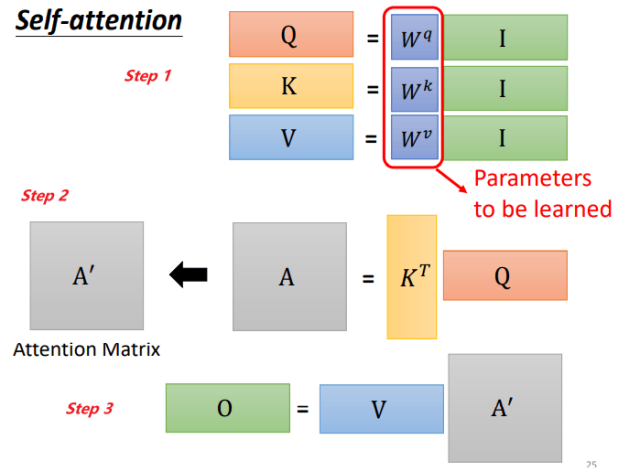

Figure 1 Overall model architecture

First, we preprocess and standardize the input life history data to ensure the quality and consistency of the data. Assume that we have a training set containing N samples, where each sample $x_i$ is a sequence data consisting of T time steps:

$$x_i = \{x_{i,1}, x_{i,2}, ..., x_{i,T}\}$$

The input $x_{i,1}$ of each time step represents the life history characteristics at a specific time point, such as eating habits, exercise frequency, sleep quality, etc. In order to enable the model to better process these time series data, we input the time series of each sample into a convolutional neural network for feature extraction.

CNN uses multiple convolutional layers to capture local features in input data. In each convolutional layer, the input data undergoes convolution operations, activation functions (such as ReLU), and pooling operations to ultimately generate a feature map. Assuming that the input data $x_i$, after the convolution operation, obtains a feature map $f_i$, it can be expressed as:

$$f_i = CNN(x_i)$$

Here, the CNN function represents the processing of the convolutional neural network. The feature map $f_i$ contains the local features extracted from the original life history data. Next, the feature map will be used as the input of the Transformer model, using the Transformer's self-attention mechanism to capture the global dependencies in the time series data.

The core of Transformer is the self-attention mechanism, which can dynamically adjust the contribution of each feature to the output by calculating the correlation between input features. Suppose we have an input sequence $f_i = \{f_{i,1}, f_{i,2}, ..., f_{i,T}\}$, where each $f_{i,t}$ represents a feature after CNN processing. In Transformer, the input sequence is first mapped to query, key, and value. The specific calculation is as follows:

$$Q_i = W_Q f_i, \quad K_i = W_K f_i, \quad V_i = W_V f_i$$

Among them, $W_Q$, $W_K$ and $W_V$ are the weight matrices of query, key and value respectively. Next, the attention weight is obtained by calculating the dot product between the query and the key, and the value is weighted summed to get the output feature.

$$z_i = Attention(Q_i, K_i, V_i) = soft\max(\frac{Q_i K_i^T}{\sqrt{d_k}}) V_i$$

Among them, $d_k$ is the dimension of the key, and the softmax function is used to normalize the calculated dot product so that the attention weight satisfies the probability distribution. Through this self-attention mechanism, the Transformer can focus on the important features in the input sequence, capture long-term dependencies, and generate global context information.

In order to enhance and upgrade the performance of the model further, we utilize a sophisticated multi-head attention mechanism. This is a state-of-the-art approach that effectively distills information from a broad array of diverse subspaces by computing several attention heads in parallel and at the same time. Accordingly, the final output features are meticulously obtained through concatenating heterogeneous outputs produced from the various attention heads and linearly transforming the combined outputs.

$$z_i^{multi-head} = Concat(z_i^1, z_i^2, ..., z_i^h) W_o$$

Where $W_o$ is the output linear transformation matrix and h is the number of attention heads. Through the multi-head attention mechanism, the model can understand the complex patterns in the input data from different angles, thereby improving the accuracy of heart disease detection.

Finally, we process the output features of Transformer through the fully connected layer to obtain the final heart disease detection results. Assuming that the feature output by the model is $z_i^{final}$, the final prediction result $y'_i$ is:

$$y'_i = \sigma(W_y z_i^{final} + b)$$

Among them, $\sigma$ is the Sigmoid activation function, which is used to map the output value to the probability range, and B and C are the weight matrix and bias term. Finally, the model output D represents the probability of the patient suffering from heart disease.

In summary, the heart disease detection algorithm based on CNN and Transformer proposed in this paper extracts local features of life history data through CNN and combines the self-attention mechanism and multi-head attention mechanism of Transformer to model global information, which can effectively improve the accuracy of heart disease prediction. This method can not only process complex time series data but also extract potential health risk signals from multi-dimensional life history information, providing strong support for the early detection of heart disease.

IV. EXPERIMENT

*A. Datasets*

The dataset used in this article is the Kaggle dataset, which contains information on multiple lifestyle and physiological indicators related to heart disease and aims to assess the risk of heart disease in individuals through these data. The columns in the dataset include basic personal information such as age, gender, place of residence, marital status, education level, and income level. In addition, it also covers a variety of factors closely related to heart health, such as blood pressure, cholesterol, diabetes, smoking, drinking, exercise level, etc. Family history and genetic risk scores of heart disease are also important features in the dataset, which can reflect the genetic tendency of individuals. By combining these variables, the model can analyze the potential impact of lifestyle habits and health status on the occurrence of heart disease.

In addition, the dataset also contains more specific physiological indicators, such as body mass index (BMI), heart rate, stress level, sleep time, etc., which may be closely related to heart health. Lifestyle factors such as medication use, hypertension, exercise, dietary fiber, and sodium intake are also taken into account. The data also provides information about individual cultural background and screen time, which may indirectly affect mental health and physical health and thus affect the risk of heart disease.

Finally, the dataset divides age into different age groups and evaluates the probability of heart disease based on factors such as personal lifestyle, work status, and family medical history. These multi-dimensional features make this dataset not only suitable for heart disease risk prediction but also provide a valuable basis for studying the influencing factors of heart disease. Through machine learning and deep learning models, we can further explore which factors are most predictive of the occurrence of heart disease and provide auxiliary support for clinical diagnosis.

*B. Experimental Results*

In order to verify the effectiveness of the proposed CNN and Transformer-based heart disease detection algorithm, we designed a series of comparative experiments. We selected a variety of classic machine learning and deep learning models as controls, covering traditional models such as logistic regression, support vector machine (SVM) [19], decision tree, and deep neural network models that have been widely used in medical data analysis in recent years, such as convolutional neural network (CNN) [20] and long short-term memory network (LSTM) [21]. Through these comparative experiments, we can not only evaluate the performance of the proposed algorithm in the task of heart disease detection but also gain an in-depth understanding of the advantages and disadvantages of different models when processing life history data. The goal of the comparative experiment is to comprehensively verify the robustness, accuracy, and ability to handle complex data of the proposed method while providing a reference for the development of future intelligent prediction systems for heart disease. By comparing with existing methods, we can clearly demonstrate the advantages of the proposed algorithm in multi-dimensional feature extraction and time series modeling, thereby highlighting its application potential in heart disease detection.

Table 1 Experimental Results

| Model | ACC | Precision | Recall |
|---|---|---|---|
| SVM | 0.75 | 0.72 | 0.78 |
| CNN | 0.81 | 0.79 | 0.83 |
| LSTM | 0.79 | 0.77 | 0.81 |
| Ours | 0.85 | 0.84 | 0.86 |

From the experimental results, the accuracy (ACC) of the SVM model is 0.75, the precision is 0.72, and the recall is 0.78, and the overall performance is relatively general. Although SVM performs well in some simple classification tasks, in the complex task of heart disease detection, its performance is limited because the model cannot effectively capture the temporal characteristics and nonlinear relationships in the data. Its low accuracy and recall also indicate that the model has certain false positives and false negatives, making it difficult to fully and accurately identify heart disease risks.

In contrast, the CNN model performs better, with an accuracy of 0.81, a precision of 0.79, and a recall of 0.83. Convolutional neural networks have significant advantages in processing image data. They effectively extract local features through convolutional layers, so they can improve the accuracy of heart disease prediction to a certain extent. However, since CNN itself does not fully consider the temporal dependence and global information in the data, although its performance is significantly better than SVM, there is still room for improvement.

The highest improvement is observed in the CNN and Transformer-based hybrid model that we have developed. The model achieved 0.85, 0.84, and 0.86 accuracy, precision, and recall rates, respectively, which are far better than other traditional models. By combining the local feature extraction capability of CNN with the global dependency modeling capability of Transformer, our model can process multi-dimensional and highly temporal information more effectively and thereby improve the accuracy and robustness of heart disease prediction. These results verify the efficacy of our proposed method on challenging datasets and also demonstrate its broad application potential in heart disease prediction.

In order to gain a deeper understanding of the contribution of each key module to the model performance, we conducted an ablation experiment to remove important components of the model one by one and observe their impact on the overall performance. The ablation experiment aims to analyze the specific role of the combination of CNN and Transformer, feature extraction, and global modeling capabilities in heart disease detection. These experiments help to reveal the key role of each module in improving prediction accuracy, precision, and recall, thereby further optimizing the model structure and improving the overall performance of the model. By gradually analyzing each part of the model, we can more clearly see which factors are most critical to improving the performance of the model, providing a theoretical basis for subsequent improvements and optimizations.

Table 2 Ablation experiment

| Model | ACC | Precision | Recall |
|---|---|---|---|
| Without Transformer | 0.81 | 0.79 | 0.83 |
| Without CNN | 0.82 | 0.83 | 0.84 |
| All | 0.85 | 0.84 | 0.86 |

From the experimental results, it can be seen that when the Transformer module is removed, the model's accuracy (ACC) is 0.81, the precision is 0.79, and the recall is 0.83. This shows that the Transformer module still plays an important role in capturing global features and processing temporal dependencies. Although the performance of the model after removal is lower than that of the complete model, it can still maintain a good effect. This shows that CNN plays a positive role in processing local features, and the Transformer can further improve the detection performance when modeling global information.

When the CNN module is removed, the performance of the model is slightly improved, with the accuracy rising to 0.82, the precision rate is 0.83, and the recall rate is 0.84, indicating that the contribution of CNN in feature extraction does not directly affect the integration of global information like Transformers. However, CNN still has an advantage in processing local features that cannot be ignored. Overall, the complete model (combining CNN and Transformer) achieved the best results, with accuracy, precision, and recall rates of 0.85, 0.84, and 0.86, respectively. This shows the complementarity of the combination of the two, which not only retains the details of local features but also effectively captures global dependencies, greatly improving the overall performance of the model.

## V. Conclusion

This paper proposes a heart disease detection algorithm based on CNN and Transformer, aiming to effectively predict the risk of heart disease from complex life history data. Through experimental verification, the proposed model shows better performance than traditional models in multiple indicators such as accuracy, precision, and recall, especially in processing multi-dimensional features and capturing global information. The results of the ablation experiment further prove the complementarity of CNN and Transformer modules. The combination of the two effectively improves the overall performance of the model, especially showing strong application potential in the task of heart disease prediction.

Although this study has achieved good results in heart disease prediction, there is still some room for improvement. For example, in the future, more external factors (such as environmental factors, genetic data, etc.) can be introduced to further enrich the input features to improve the generalization ability of the model. In addition, the training process of the model can explore more efficient optimization strategies, such as adaptive learning rate adjustment and more sophisticated feature selection methods, to further improve the training speed and performance.

Looking forward to the future, with the continuous advancement of big data and artificial intelligence technology, intelligent medical systems based on machine learning and deep learning will be applied in more fields, especially in early disease prediction and the formulation of personalized medical plans. Future research can further expand the application scenarios of this model, combine more physiological data and clinical history, and develop more accurate and efficient early screening tools for heart disease, providing strong support for achieving more extensive health management.